\pdfoutput=1
\documentclass[11pt,letterpaper]{article}
\usepackage{naaclhlt2016}
\usepackage{times}
\usepackage{latexsym}
\usepackage{microtype}
\usepackage{latexsym}
\usepackage{array,arydshln}
\usepackage{multirow}
\usepackage{graphics}
\usepackage{pgfplots}
\usepackage{float} 
\usepackage{tikz}
\usepackage{amsmath}
\usepackage{amssymb}
\usepackage{amsfonts}
\usepackage{relsize}
\usepackage{multirow}
\usepackage{url}
\usepackage{array}
\usepackage[font=small]{caption}

\pgfplotsset{every tick label/.append style={font=\small}}

\naaclfinalcopy 
\def\naaclpaperid{100} 

\title{Counter-fitting Word Vectors to Linguistic Constraints} 

\author{Nikola Mrk\v{s}i\'c$^{\mathbf{1}}$ ~ Diarmuid {\'O S\'eaghdha}$^{\mathbf{2}}$ ~ Blaise Thomson$^{\mathbf{2}}$ ~ Steve Young$^{\mathbf{1}}$  \\
 $^{\mathbf{1}}$  Department of Engineering, University of Cambridge, UK  \\
  $^{\mathbf{2}}$ Apple Inc.  \\
  {\small \tt \{nm480, sjy\}@eng.cam.ac.uk} \\ {\small \tt\{doseaghdha, blaisethom\}@apple.com}}

\fi

\author{Nikola Mrk\v{s}i\'c$^{\mathbf{1}}$, ~ Diarmuid {\'O S\'eaghdha}$^{\mathbf{2}}$, ~ Blaise Thomson$^{\mathbf{2}}$, ~ Milica Ga\v{s}i\'c$^{\mathbf{1}}$   \\ \bf  Lina Rojas-Barahona$^{\mathbf{1}}$, ~Pei-Hao Su$^{\mathbf{1}}$, ~David Vandyke$^{\mathbf{1}}$, ~Tsung-Hsien Wen$^{\mathbf{1}}$, ~Steve Young$^{\mathbf{1}}$  \\
 $^{\mathbf{1}}$  Department of Engineering, University of Cambridge, UK  \\
  $^{\mathbf{2}}$ Apple Inc.  \\
  {\small \tt \{nm480,mg436,phs26,djv27,thw28,sjy\}@cam.ac.uk} \\ {\small \tt\{doseaghdha, blaisethom\}@apple.com}}

\begin{document}
\maketitle
\begin{abstract} 
In this work, we present a novel \emph{counter-fitting} method which injects antonymy and synonymy constraints into vector space representations in order to improve the vectors' capability for judging semantic similarity. Applying this method to publicly available pre-trained word vectors leads to a new state of the art performance on the SimLex-999 dataset. We also show how the method can be used to tailor the word vector space for the downstream task of dialogue state tracking, resulting in robust improvements across different dialogue domains. 
\end{abstract}

\section{Introduction}

Many popular methods that induce representations for words rely on the \emph{distributional hypothesis} -- the assumption that semantically similar or related words appear in similar contexts. This hypothesis supports unsupervised learning of meaningful word representations from large corpora \cite{Curran:03,OSeaghdha:Korhonen:14,Mikolov:13,Pennington:14}. Word vectors trained using these methods have proven useful for many downstream tasks including machine translation \cite{emnlp:mt:2013} and dependency parsing \cite{bansal:2014}.

One drawback of learning word embeddings from co-occurrence information in corpora is that it tends to coalesce the notions of \emph{semantic similarity} and \emph{conceptual association} \cite{Hill:2014}. Furthermore, even methods that can distinguish similarity from association (e.g., based on syntactic co-occurrences) will generally fail to tell synonyms from antonyms \cite{Mohammad:EtAl:08}. For example, words such as \emph{east} and \emph{west} or \emph{expensive} and \emph{inexpensive} appear in near-identical contexts, which means that distributional models produce very similar word vectors for such words. Examples of such anomalies in GloVe vectors can be seen in Table \ref{tab:neighbours}, where words such as \emph{cheaper} and \emph{inexpensive} are deemed similar to (their antonym) \emph{expensive}. 

\begin{table}
\centering
\begin{tabular}{c|ccc}

&\bf east & \bf expensive & \bf British \\ \hline 

\multirow{5}{*}{Before} & west & pricey & American \\
&north & cheaper & Australian \\
&south & costly & Britain \\
&~southeast~ & ~overpriced~ & ~European~ \\
&~northeast~ & ~inexpensive~ & England \\

\hline 
\multirow{5}{*}{After} & eastward & costly & Brits \\
& eastern & pricy & London \\
& easterly & overpriced & BBC\\
& - & pricey & UK \\
& - & afford & Britain  

\end{tabular}%
\caption{Nearest neighbours for target words using GloVe vectors before and after counter-fitting}\label{tab:neighbours}
\end{table}

A second drawback is that similarity and antonymy can be application- or domain-specific. In our case, we are interested in exploiting distributional knowledge for the \emph{dialogue state tracking} task (DST). The DST component of a dialogue system is responsible for interpreting users' utterances and updating the system's \textit{belief state} -- a probability distribution over all possible states of the dialogue. For example, a DST for the restaurant domain needs to detect whether the user wants a \textit{cheap} or \textit{expensive} restaurant. Being able to generalise using distributional information while still distinguishing between semantically different yet conceptually related words (e.g.~\emph{cheaper} and \emph{pricey}) is critical for the performance of dialogue systems. In particular, a dialogue system can be led seriously astray by false synonyms.

We propose a method that addresses these two drawbacks by using synonymy and antonymy relations drawn from either a general lexical resource or an application-specific ontology to fine-tune distributional word vectors. Our method, which we term \emph{counter-fitting}, is a lightweight post-processing procedure in the spirit of \emph{retrofitting} \cite{faruqui:15}. The second row of Table \ref{tab:neighbours} illustrates the results of counter-fitting: the nearest neighbours capture true similarity much more intuitively than the original GloVe vectors.
The procedure improves word vector quality regardless of the initial word vectors provided as input.\footnote{When we write ``improve'', we refer to improving the vector space for a specific purpose. We do not expect that a vector space fine-tuned for semantic similarity will give better results on semantic relatedness. As Mohammad et al.~\shortcite{Mohammad:EtAl:08} observe, antonymous concepts are  related but not similar.}
By applying counter-fitting to the Paragram-SL999 word vectors provided by Wieting et al.~\shortcite{Wieting:15}, we achieve new state-of-the-art performance on SimLex-999, a dataset designed to measure how well different models judge semantic similarity between words \cite{Hill:2014}.
We also show that the counter-fitting method can inject knowledge of dialogue domain ontologies into word vector space representations to facilitate the construction of semantic dictionaries which improve DST performance across two different dialogue domains. Our tool and word vectors are available at \url{github.com/nmrksic/counter-fitting}.

\section{Related Work} 

Most work on improving word vector representations using lexical resources has focused on bringing words which are known to be semantically related closer together in the vector space. Some methods modify the prior or the regularization of the original training procedure \cite{Yu:2014,Bian:14,kiela:15}. Wieting et al.~\shortcite{Wieting:15} use the Paraphrase Database \cite{ppdb:13} to train word vectors which emphasise word similarity over word relatedness. These word vectors achieve the current state-of-the-art performance on the SimLex-999 dataset and are used as input for counter-fitting in our experiments. 

Recently, there has been interest in lightweight post-processing procedures that use lexical knowledge to refine off-the-shelf word vectors without requiring large corpora for (re-)training as the aforementioned ``heavyweight" procedures do. Faruqui et al.'s \shortcite{faruqui:15} \textit{retrofitting} approach uses similarity constraints from WordNet and other resources to pull similar words closer together.

The complications caused by antonymy for distributional methods are well-known in the semantics community. Most prior work focuses on extracting antonym pairs from text rather than exploiting them \cite{Lin:EtAl:03,Mohammad:EtAl:08,Turney:08,Hashimoto:EtAl:12,Mohammad:EtAl:13}. The most common use of antonymy information is to provide features for systems that detect contradictions or logical entailment \cite{Marcu:Echihabi:02,DeMarneffe:EtAl:08,Zanzotto:EtAl:09}. As far as we are aware, there is no previous work on exploiting antonymy in dialogue systems. The modelling work closest to ours are Liu et al.~\shortcite{Liu:EtAl:15}, who use antonymy and WordNet hierarchy information to modify the heavyweight Word2Vec training objective; Yih et al.~\shortcite{Yih:EtAl:12}, who use a Siamese neural network to improve the quality of 
Latent Semantic Analysis vectors; Schwartz et al.~\shortcite{schwartz-reichart-rappoport:2015:Conll}, who build a standard distributional model from co-occurrences based on \textit{symmetric patterns}, with specified antonymy patterns counted as negative co-occurrences; and Ono et al.~\shortcite{Ono:15}, who use thesauri and distributional data to train word embeddings specialised for capturing antonymy.

\section{Counter-fitting Word Vectors to Linguistic Constraints}
Our starting point is an indexed set of word vectors $V = \left\{ \mathbf{v}_{1}, \mathbf{v}_{2}, \ldots, \mathbf{v}_{N} \right\}$ with one vector for each word in the vocabulary. We will inject semantic relations into this vector space to produce new word vectors $V' = \left\{ \mathbf{v'}_{1}, \mathbf{v'}_{2}, \ldots, \mathbf{v'}_{N} \right\} $. For antonymy and synonymy we have a set of constraints $A$ and $S$, respectively. The elements of each set are pairs of word indices; for example, each pair $(i,j)$ in $S$ is such that the $i$-th and $j$-th words in the vocabulary are synonyms. The objective function used to {counter-fit} the pre-trained word vectors $V$ to the sets of linguistic constraints ${A}$ and ${S}$ contains three different terms:

 \textbf{1. Antonym Repel ({AR})}:  This term serves to \emph{push} antonymous words' vectors away from each other in the transformed vector space $V'$: 
\begin{equation}
\text{AR}(V') = \sum_{(u, w) \in A} \tau\left( \delta - d( \mathbf{v}_{u}', \mathbf{v}_{w}' ) \right)\nonumber
\end{equation}
\noindent where $d(v_i,v_j) = 1 - \operatorname{cos}(v_i,v_j)$ is a distance derived from cosine similarity and $\tau(x) \triangleq {max}(0, x)$ imposes a margin on the cost. Intuitively, $\delta$ is the ``ideal'' minimum distance between antonymous words; in our experiments we set $\delta = 1.0$ as it corresponds to vector orthogonality. 

\vspace{2mm}

\textbf{2. Synonym Attract (SA)}: The counter-fitting procedure should seek to bring the word vectors of known synonymous word pairs closer together:
\begin{equation}
\text{SA}(V') = \sum_{(u, w) \in S} \tau\left( d(\mathbf{v}_{u}', \mathbf{v}_{w}') - \gamma\right)\nonumber
\end{equation}
\noindent where $\gamma$ is the ``ideal'' maximum distance between synonymous words; we use $\gamma= 0$. 

\vspace{2mm}

\textbf{3. Vector Space Preservation (VSP)}: the topology of the original vector space describes relationships between words in the vocabulary captured using distributional information from very large textual corpora. The VSP term bends the transformed vector space towards the original one as much as possible in order to preserve the semantic information contained in the original vectors:
\begin{equation}
\text{VSP}(V, V') = \sum_{i = 1 }^{N} \sum_{j \in {N}(i)} \tau\left( d(\mathbf{v}_{i}', \mathbf{v}_{j}') - d(\mathbf{v}_i, \mathbf{v}_j)\right)\nonumber
\end{equation}
\noindent For computational efficiency, we do not calculate distances for every pair of words in the vocabulary. Instead, we focus on the (pre-computed) neighbourhood ${N}(i)$, which denotes the set of words within a certain radius $\rho$ around the $i$-th word's vector in the original vector space $V$. Our experiments indicate that counter-fitting is relatively insensitive to the choice of $\rho$, with values between 0.2 and 0.4 showing little difference in quality; here we use $\rho = 0.2$. \vspace{1mm}

The objective function for the training procedure is given by a weighted sum of the three terms:
\begin{equation}
C(V, V') = k_1 \text{AR}(V') + k_2 \text{SA}(V') + k_3 \text{VSP}(V,V')\nonumber
\end{equation}
\noindent where $k_1, k_2, k_3 \ge 0$ are hyperparameters that control the relative importance of each term. In our experiments we set them to be equal: $k_1 = k_2 = k_3$. To minimise the cost function for a set of starting vectors $V$ and produce counter-fitted vectors $V'$, we run stochastic gradient descent (SGD) for 20 epochs. 
An end-to-end run of counter-fitting takes less than two minutes on a laptop with four CPUs.

\subsection{Injecting Dialogue Domain Ontologies into Vector Space Representations}

Dialogue state tracking (DST) models capture users' goals given their utterances. Goals are represented as sets of constraints expressed by \emph{slot-value} pairs such as [food: \emph{Indian}] or [parking: \emph{allowed}]. The set of slots $S$ and the set of values $V_s$ for each slot make up the \textit{ontology} of a dialogue domain. 

In this paper we adopt the recurrent neural network (RNN) framework for tracking suggested in \cite{Henderson:14b,Henderson:14d,Mrksic:15}. Rather than using a spoken language understanding (SLU) decoder to convert user utterances into meaning representations, this model operates directly on the $n$-gram features extracted from the automated speech recognition (ASR) hypotheses. A drawback of this approach is that the RNN model can only perform exact string matching to detect the slot names and values mentioned by the user. It cannot recognise synonymous words such as \emph{pricey} and \emph{expensive}, or even subtle morphological variations such as \emph{moderate} and \emph{moderately}. A simple way to mitigate this problem is to use \emph{semantic dictionaries}: lists of rephrasings for the values in the ontology. Manual construction of dictionaries is highly labour-intensive; however, if one could automatically detect high-quality rephrasings, then this capability would come at no extra cost to the system designer.

To obtain a set of word vectors which can be used for creating a semantic dictionary, we need to \emph{inject} the domain ontology into the vector space. This can be achieved by introducing antonymy constraints between all the possible values of each slot (i.e.~\emph{Chinese} and \emph{Indian}, \emph{expensive} and \emph{cheap}, etc.). The remaining linguistic constraints can come from semantic lexicons: the richer the sets of injected synonyms and antonyms are, the better the resulting word representations will become.

\section{Experiments}

 \subsection{Word Vectors and Semantic Lexicons}

 Two different collections of pre-trained word vectors were used as input to the counter-fitting procedure:
\begin{enumerate}
\item \textbf{ Glove Common Crawl} 300-dimensional vectors made available by Pennington et al.~\shortcite{Pennington:14}. 
\item \textbf{ Paragram-SL999} 300-dimensional vectors made available by Wieting et al.~\shortcite{Wieting:15}.
\end{enumerate}

The synonymy and antonymy constraints were obtained from two semantic lexicons: 

\begin{enumerate}
\item  \textbf{PPDB 2.0 \cite{PPDB2}:} the latest release of the Paraphrase Database. A new feature of this version is that it assigns relation types to its word pairs. We identify the \emph{Equivalence} relation with synonymy and \emph{Exclusion} with antonymy. We used the largest available (XXXL) version of the database and only considered single-token terms.

\item \textbf{WordNet \cite{Miller:95}}: a well known semantic lexicon which contains vast amounts of high quality human-annotated synonym and antonym pairs. Any two words in our vocabulary which had antonymous word senses were considered antonyms; WordNet synonyms were not used. 

\end{enumerate}


\noindent In total, the lexicons yielded 12,802 antonymy and 31,828 synonymy pairs for our vocabulary, which consisted of 76,427 most frequent words in OpenSubtitles, obtained from \url{invokeit.wordpress.com/frequency-word-lists/}.

\begin{table} [t]
\begin{center}
\resizebox{1.0\columnwidth}{!}{%
\begin{tabular}{|l|c|}
\hline
 \bf Model / Word Vectors & $\rho$ \\ \hline 
 Neural MT Model \cite{Hill:14a} & 0.52 \\ 
 Symmetric Patterns \cite{schwartz-reichart-rappoport:2015:Conll} & 0.56  \\ 
 Non-distributional Vectors \cite{faruqui:2015b} & 0.58 \\ 

GloVe vectors \cite{Pennington:14} & 0.41  \\ 
GloVe vectors + Retrofitting & 0.53 \\ 

GloVe + Counter-fitting &  0.58   \\ 
Paragram-SL999 \cite{Wieting:15}   & 0.69 \\
Paragram-SL999 + Retrofitting & 0.68 \\  
Paragram-SL999 + Counter-fitting & \textbf{0.74} \\  \hline
  Inter-annotator agreement & 0.67 \\  
 Annotator/gold standard agreement & 0.78 \\ \hline

\end{tabular}%
}%
	\caption{Performance on SimLex-999. Retrofitting uses the code and (PPDB) data provided by the authors \vspace{-1mm}
  }\label{tab:simlex-scores}
\end{center}
\end{table}

\subsection{Improving Lexical Similarity Predictions}

In this section, we show that counter-fitting pre-trained word vectors with linguistic constraints improves their usefulness for judging semantic similarity. We use Spearman's rank correlation coefficient with the SimLex-999 dataset, which contains word pairs ranked by a large number of annotators instructed to consider only semantic similarity. 

Table \ref{tab:simlex-scores} contains a summary of recently reported competitive scores for SimLex-999, as well as the performance of the {unaltered}, {retrofitted} and {counter-fitted} GloVe and Paragram-SL999 word vectors. To the best of our knowledge, the 0.685 figure reported for the latter represents the current high score. This figure is above the average inter-annotator agreement of 0.67, which has been referred to as the ceiling performance in most work up to now. 

In our opinion, the average inter-annotator agreement is not the only meaningful measure of ceiling performance. We believe it also makes sense to compare: \textbf{a)} the model ranking's correlation with the gold standard ranking to: \textbf{b)} the average rank correlation that individual human annotators' rankings achieved with the gold standard ranking. The SimLex-999 authors have informed us that the average annotator agreement with the gold standard is 0.78.\footnote{This figure is now reported as a potentially fairer ceiling performance on the SimLex-999 website: \url{http://www.cl.cam.ac.uk/~fh295/simlex.html}.} As shown in Table \ref{tab:simlex-scores}, the reported performance of all the models and word vectors falls well below this figure.

\begin{table} 
\resizebox{1.0\columnwidth}{!}{%
\begin{tabular}{| l | c | c |}
\hline
\bf Semantic Resource & { Glove } & { Paragram } \\ \hline 
Baseline (no linguistic constraints) & 0.41 & 0.69 \\ \hline
PPDB$-$ (PPDB antonyms)  & 0.43 &  0.69 \\
PPDB$+$ (PPDB synonyms) &  0.46 & 0.68 \\ 
WordNet$-$ (WordNet antonyms) & 0.52 & 0.74 \\ \hline
PPDB$-$ and PPDB$+$ & 0.50 & 0.69 \\
WordNet$-$ and PPDB$-$ & 0.53 &  0.74  \\
WordNet$-$ and PPDB$+$ & 0.58  & 0.74 \\ \hline
WordNet$-$ and PPDB$-$ and PPDB$+$ & 0.58 & 0.74 \\ \hline
\end{tabular}%
}%
\caption{SimLex-999 performance when different sets of linguistic constraints are used for counter-fitting}\label{tab:simlex-scores-variation}\vspace{-0mm} \end{table}%

Retrofitting pre-trained word vectors improves GloVe vectors, but not the already semantically specialised Paragram-SL999 vectors. Counter-fitting substantially improves both sets of vectors, showing that injecting antonymy relations goes a long way towards improving word vectors for the purpose of making semantic similarity judgements.

Table \ref{tab:simlex-scores-variation} shows the effect of injecting different categories of linguistic constraints. GloVe vectors benefit from all three sets of constraints, whereas the quality of Paragram vectors, already exposed to PPDB, only improves with the injection of WordNet antonyms. 
Table \ref{tab:simlex-errors} illustrates how incorrect similarity predictions based on the original (Paragram) vectors can be fixed through counter-fitting. The table presents eight false synonyms and nine false antonyms: word pairs with predicted rank in the top (bottom) 200 word pairs and gold standard rank 500 or more positions lower (higher). Eight of these errors are fixed by counter-fitting: the difference between predicted and gold-standard ranks is now 100 or less. Interestingly, five of the eight corrected word pairs do not appear in the sets of linguistic constraints; these are indicated by double ticks in the table. This shows that {secondary (i.e.~\emph{indirect}) interactions} through the three terms of the cost function do contribute to the semantic content of the transformed vector space.

 \begin{table} 
 \centering
 \resizebox{1\columnwidth}{!}{%
 \begin{tabular}{| c | c | c | c |}
 \hline

 \bf  {False Synonyms} & \bf Fixed & \bf {False Antonyms} & \bf Fixed \\ \hline 


 \small{sunset, sunrise} & \checkmark & \small{dumb, dense} &   \\ 
 \small{forget, ignore} &  &\small{adult, guardian} & \\ 
 \small{girl, maid} & & \small{polite, proper} & \checkmark \checkmark \\ 
 \small{happiness, luck} & \checkmark \checkmark & \small{strength, might} &  \\ 
 \small{south, north} & \checkmark & \small{water, ice}&\\ 
 \small{go, come} & \checkmark & \small{violent, angry}& \checkmark \checkmark \\ 
 \small{groom, bride} & & \small{cat, lion}& \checkmark \checkmark \\ 
 \small{dinner, breakfast} & & \small{laden, heavy}& \checkmark \checkmark \\ 
 \small{-} & \small{-} &\small{engage, marry}&   \\ \hline

 \end{tabular}%
 }%
 \caption{Highest-error SimLex-999 word pairs using Paragram vectors (before counter-fitting)\vspace{-3mm}}\label{tab:simlex-errors}
 \end{table}%

\subsection{Improving Dialogue State Tracking}

Table \ref{tab:datasets} shows the dialogue state tracking datasets used for evaluation. These datasets come from the Dialogue State Tracking Challenges 2 and 3 \cite{Henderson:14a,Henderson:14c}. 

We used four different sets of word vectors to construct semantic dictionaries: the original GloVe and Paragram-SL999 vectors, as well as versions counter-fitted to each domain ontology. The constraints used for counter-fitting were all those from the previous section as well as antonymy constraints among the set of values for each slot. We treated all vocabulary words within some radius $t$ of a slot value as rephrasings of that value. The optimal value of $t$ was determined using a grid search: we generated a dictionary and trained a model for each potential $t$, then evaluated on the development set. Table \ref{tab:semdict-lex-rich-ru} shows the performance of RNN models which used the constructed dictionaries. The dictionaries induced from the pre-trained vectors substantially improved tracking performance over the baselines (which used no semantic dictionaries). The dictionaries created using the counter-fitted vectors improved performance even further. Contrary to the SimLex-999 experiments, starting from the Paragram vectors did not lead to superior performance, which shows that injecting the application-specific ontology is at least as important as the quality of the initial word vectors.

\begin{table} [t]
\resizebox{1.0\columnwidth}{!}{%
\begin{tabular}{|c|cccc|}
\hline
\bf Dataset & \bf Train & \bf Dev & \bf Test & \bf \#Slots \\ \hline
\bf Restaurants &  1612 & 506 & 1117 & 4 \\ 
\bf Tourist Information &  1600 & 439 &  225 & 9 \\ \hline
\end{tabular}%
}%
\caption{\label{tab:datasets} Number of dialogues in the dataset splits used for the Dialogue State Tracking experiments}
\end{table}%

\begin{table} [t]
\resizebox{1.0\columnwidth}{!}{%
\begin{tabular}{|l|c|c|}
\hline
\bf Word Vector Space & \bf Restaurants & \bf Tourist Info    \\ \hline
 Baseline (no dictionary)  & 68.6 &  60.5  \\  \hline
 GloVe & 72.5 & 60.9 \\
 GloVe + Counter-fitting & 73.4 & {\bf 62.8}	 \\ 
 Paragram-SL999 & 73.2 & 61.5 \\ 	
 Paragram-SL999 + Counter-fitting & {\bf 73.5} & 61.9 \\ \hline
\end{tabular}%
}%
\caption{\label{tab:semdict-lex-rich-ru} Performance of RNN belief trackers (ensembles of four models) with different semantic dictionaries \vspace{-1mm}
}
\end{table}%

\section{Conclusion}

We have presented a novel \emph{counter-fitting} method for injecting linguistic constraints into word vector space representations. The method efficiently post-processes word vectors to improve their usefulness for tasks which involve making semantic similarity judgements. Its focus on separating vector representations of antonymous word pairs lead to substantial improvements on genuine similarity estimation tasks. We have also shown that counter-fitting can tailor word vectors for downstream tasks by using it to inject domain ontologies into word vectors used to construct semantic dictionaries for dialogue systems. 

\section*{Acknowledgements}		

We would like to thank Felix Hill for help with the SimLex-999 evaluation. We also thank the anonymous reviewers for their helpful suggestions.		
\clearpage

\bibliography{naaclhlt2016}
\bibliographystyle{naaclhlt2016}

\end{document}